\pdfoutput=1
\documentclass[]{interact}

\usepackage{epstopdf}
\usepackage{subfigure}

\usepackage{natbib}
\bibpunct[, ]{(}{)}{;}{a}{}{,}
\renewcommand\bibfont{\fontsize{10}{12}\selectfont}

\usepackage{amssymb}
\usepackage{adjustbox}
\usepackage{float}
\usepackage{color,soul}
\usepackage{listings}
\usepackage{xcolor}

\restylefloat{table}

\definecolor{codebkg}{HTML}{EBEBEB}

\newcommand{\orcidauthorJMR}{0000-0002-3665-639X}
\newcommand{\orcidauthorJL}{0000-0001-5044-2921}
\newcommand{\orcidauthorDM}{0000-0003-4480-082X}
\newcommand{\orcidauthorBF}{0000-0002-8585-9388}
\newcommand{\orcidauthorDK}{0000-0003-3660-9187}
\newcommand{\orcidauthorzxc}{0000-0003-1506-3314}

\begin{document}


\title{Actionable Cognitive Twins for Decision Making in Manufacturing}

\author{
\name{Jo\v{z}e M. Ro\v{z}anec\textsuperscript{a,b,c}\orcidauthorJMR{}, Jinzhi Lu\textsuperscript{d}\thanks{CONTACT Jinzhi Lu. Email: jinzhl@epfl.ch}\orcidauthorJL{}, Jan Rupnik\textsuperscript{b,c}, Maja \v{S}krjanc\textsuperscript{b,c}, Dunja Mladeni\'{c}\textsuperscript{b}\orcidauthorDM{}, Bla\v{z} Fortuna\textsuperscript{b,c}\orcidauthorBF{}, Xiaochen Zheng\textsuperscript{d}\orcidauthorzxc{} and Dimitris Kiritsis\textsuperscript{d}\orcidauthorDK{}}
\affil{\textsuperscript{a}Jo\v{z}ef Stefan International Postgraduate School, Jamova 39, 1000 Ljubljana, Slovenia; \textsuperscript{b}Jo\v{z}ef Stefan Institute, Jamova 39, 1000 Ljubljana, Slovenia; \textsuperscript{c}Qlector d.o.o., Rov\v{s}nikova 7, 1000 Ljubljana, Slovenia; \textsuperscript{d}EPFL SCI-STI-DK, Station 9, CH-1015 Lausanne, Switzerland}
}

\maketitle

\begin{abstract}
Actionable Cognitive Twins are the next generation Digital Twins enhanced with cognitive capabilities through a knowledge graph and artificial intelligence models that provide insights and decision-making options to the users. The knowledge graph describes the domain-specific knowledge regarding entities and interrelationships related to a manufacturing setting. It also contains information on possible decision-making options that can assist decision-makers, such as planners or logisticians. In this paper, we propose a knowledge graph modeling approach to construct actionable cognitive twins for capturing specific knowledge related to demand forecasting and production planning in a manufacturing plant. The knowledge graph provides semantic descriptions and contextualization of the production lines and processes, including data identification and simulation or artificial intelligence algorithms and forecasts used to support them. Such semantics provide ground for inferencing, relating different knowledge types: creative, deductive, definitional, and inductive. To develop the knowledge graph models for describing the use case completely, systems thinking approach is proposed to design and verify the ontology, develop a knowledge graph and build an actionable cognitive twin. Finally, we evaluate our approach in two use cases developed for a European original equipment manufacturer related to the automotive industry as part of the European Horizon 2020 project FACTLOG.
\end{abstract}


\section{Introduction}\label{INTRODUCTION}


Digitization of manufacturing has become a requirement for manufacturing companies. It provides an approach to increase control capabilities and transparency of production processes with a high level of scalability; it accelerates the information flow within the organizations and sets up the base for innovation of manufactured products at an organizational level. Digital Twins (DTs) are considered as core to support the digitization of manufacturing processes. \cite{shafto2010draft} first proposed Digital Twins' concept within the NASA Technology Roadmap to support complex system development. Since then, multiple understandings of the Digital Twins concept were developed. New variations enable different views of complex system development. Regardless of the emphasis, a common understanding of the Digital Twin is defined as a digital counterpart of a physical reality used to analyze and improve its performance and "derive solutions relevant for the real system" (\cite{boschert2018next}).

To realize this purpose, multiple authors envisioned the inclusion of semantic technologies into the Digital Twins to describe the reference components, their relationships, and how they relate to their digital counterparts. For example, \cite{banerjee2017generating} developed an ontology to construct data pipelines that allow extracting semantic relations from sensor data and plant features described by a Knowledge Graph (KG). By providing semantics to the data flow, required features can be captured to support decision-making by querying the Knowledge Graph. \cite{rozanec2020sedit} envisioned using ontologies to model reference entities in Digital Twins' context and implement decision-making for production process management and anomaly analysis. \cite{zehnder2018representing} made use of semantics modeling to formalize the data channels between Digital Twins. By identifying data sources as labeled virtual sensors semantically, they provide relevant information on data such as its characteristics and usage. \cite{gomez2019sedit} provided a semantic model for automatic discovery of Digital Twins and manage interoperability between Digital Twins. The value derived from semantic integration of Digital Twins was also recognized by \cite{boschert2018next}. In the \textit{NextDT} paradigm, they propose a solution connecting Digital Twins as a value network to orchestrate multiple services. \cite{JINZHIct} proposed using ontologies and timestamps to model the evolution of entities in time, providing ontology-based topologies representing transitions between already known and modeled states.

Though authors envisioned the usage of semantic technologies in Digital Twins' context to manage data and services, they barely addressed the dimension of cognition in Digital Twins and semantic technologies' role in it. We thus further develop the concept of Actionable Cognitive Twin (ACT) in the context of manufacturing introduced by \cite{rozanec2020sedit}, focusing on the cognitive aspect while also deepening the understanding of the actionable dimension. ACTs cognitive dimension requires a semantic mapping of entities and their relationships, artificial intelligence (AI) and statistical models that provide inductive knowledge, and a set of rules that enable reasoning. On the actionable dimension, ACTs display some interaction, can trigger actions, and assist in decision-making by providing contextualized decision-making options to end-users, who shall make a decision.

This paper provides several original contributions. First, we evolve the Actionable Cognitive Twin definition presented in \cite{rozanec2020sedit}, emphasizing the cognitive and actionable dimensions in a manufacturing setting. Second, a unified ontology is proposed to capture domain knowledge regarding manufacturing plants, use cases, and integrate Digital Twin components with AI and statistical models, predictions, explanations, and decision-making options. Third, we provide a system thinking approach to define the ontology and construct the Knowledge Graph based on it and best practices outlined in \cite{schindler2019building,schwabe2019building,liebig2019building}. Finally, we evaluate the Actionable Cognitive Twin on two use cases: (Demand Forecasting (DF) and Production Planning (PP)), using data provided by manufacturing partners of the European Horizon 2020 project FACTLOG.

To evaluate the Actionable Cognitive Twin concept, we provide sample questions for each use case, which can not be resolved in a Digital Twin setting but are solved with an Actionable Cognitive Twin. We translate those questions to queries, which we execute against the Knowledge Graph. The queries showcase how different types of knowledge can be related and how decision-making options can be provided for a given use case or entity of a manufacturing setting. Finally, we also perform a quantitative evaluation of the Knowledge Graph through graph metrics.

The rest of this paper is structured as follows: Section~\ref{RELATED-WORK} presents related work, Section~\ref{ACT-DEFINITION} evolves the concept of Actionable Cognitive Twins in a manufacturing context, Section~\ref{SYSTEM-THINKING} describes the system thinking approach we used to construct Actionable Cognitive Twins, Section~\ref{USE-CASES} describes the two use cases we considered to build the ACT, Section~\ref{RESULTS} provides the results we obtained on evaluating the Actionable Cognitive Twin. Finally, in Section~\ref{CONCLUSION}, we provide our conclusions and outline future work.

\section{Related Work}\label{RELATED-WORK}

\subsection{Digital Twins in manufacturing}

Digital Twin's concept was first proposed by \cite{shafto2010draft} within the NASA Technology Roadmap to support complex system development. The Digital Twin can be defined as a whole comprised of a reference asset and its digital representation that evolve together and are connected by a bidirectional data and information flow (\cite{sjarov2020digital, qi2019enabling}). Digital Twins inform on multiple aspects of the entity in a consistent format (\cite{kritzinger2018digital}), which can be consumed by other services or assist people in decision-making. Specifically, some researchers may include a semantic data model to describe relationships between elements (\cite{kunath2018integrating}). Digital Twins are not just data placeholders: they interact with other Digital Twins and the physical system, providing insights and value that transcends ingested data. Such ability to provide insights and value is considered by \cite{grieves2017digital} as Digital Twins behavior. Such behavior is achieved through software analytics, AI, and statistical models (\cite{bevilacqua2020digital}). These approaches support static and dynamic analysis, propagate the information to related Digital Twins, and act promptly on detected variations. To facilitate analysis and information propagation in complex systems, Digital Twins can be composed hierarchically, by association or peer-to-peer (\cite{becue2020new}), according to the natural relationship of components.

To mirror the physical object or process, Digital Twins in manufacturing require extracting data from multiple sources to provide consistent information for decision making (\cite{lu2020digital,damjanovic2019open}). First, in order to realize the unified format across different Digital Twins, researchers make use of semantic models to assign data to semantic entities (\cite{damjanovic2019open,olivotti2019creating,schluse2018experimentable}), to represent components and their relationships (\cite{rosen2015importance,bao2019modelling,catarci2019conceptual}). Second, Digital Twins in manufacturing display behavior (e.g.: inform most likely outcome of a workflow) by reacting to inputs achieved through simulations (\cite{lu2020digital,rosen2015importance,papanagnou2020digital,shao2018digital}) (e.g.: workflow simulations) and AI models (\cite{post2017physical,jaensch2018digital}) (e.g.: demand forecasting models) to provide a realistic insight on likely future outcomes. E.g., Monte Carlo simulations can be used to generate production plans and schedules based on data distributions observed in the past to provide probabilistic estimates of future outcomes. Similarly, AI models can learn from data of past demand and learn most likely future outcomes. Finally, Digital Twins integration into more complex systems was recognized by several authors as means to deal with complex problems and seek solutions (\cite{bao2019modelling,abburu2020cognitwin}). For example, in a shop-floor setting, Digital Twins are developed at different levels. Their integration mirrors interactions between people and machines, machines and objects, or between machines.

To promote understanding of complex systems and support their operations, cognitive and actionable capabilities are required on Digital Twins. Digital Twins cognitive dimension in manufacturing was addressed by \cite{JINZHIct}, stressing the importance of a Knowledge Graph to be able to model entities' evolution in time through ontology topologies and timestamps. \cite{eirinakis2020enhancing} described a different approach. The cognition process is triggered by Digital Twin inputs and comprehends reasoning, supported by just-in-time simulations. We agree with the authors and extend the vision on cognition to create deductive and inductive knowledge and relate it to definitional knowledge already encoded in the Digital Twin. Digital Twins actionable dimension was first mentioned by \cite{grieves2017digital}, understanding it as Digital Twins simulation capabilities. \cite{rozanec2020sedit} extended this understanding in the context of manufacturing to the capability to inform the user about decision-making opportunities. We agree on both definitions and consider the actionable dimension as the ability to produce knowledge and interact with the environment informing it, proposing decision-making opportunities, or taking action where possible.

Currently, Digital Twins describe their building parts and process data for specific domains. However, there is a huge challenge to describe domain-specific knowledge with a unified format, especially when integrating heterogeneous Digital Twins to model a complex system. This can be achieved through a unified ontology to support the manufacturing use cases for which the ACT is built. Moreover, to promote understanding of complex systems and support their operations, cognitive and actionable capabilities need to be enhanced on Digital Twins. Cognition capabilities enhancement would enable better integration of domain-specific knowledge and provide better cues to real-time decision-making.

\subsection{Domain specific ontologies}\label{RELATED-WORK-DSO}
Demand Forecasting and Production Planning are two important activities in manufacturing. Demand Forecasting tries to provide an accurate estimate of future demand and associated uncertainty to inform planners' decisions. Production Planning is concerned with the efficient resource allocation to produce the required amount of goods meeting expected deadlines. They are closely related to each other: demand forecasts inform the planner about product quantities expected for future dates to order required raw materials, schedule workers and production lines, and anticipate any issues that can prevent in-time delivery.

Many domain ontologies capture concepts of Demand Forecasting and Production Planning. Some Production Planning concepts can be found in the Manufacturing Service Description Language (MSDL) ontology (\cite{ameri2006upper,ameri2012systematic}), which primarily provides a formal representation of manufacturing services in the mechanical machining domain. Production knowledge regarding individual operations is described in MASON (\cite{lemaignan2006mason}). Elements useful to both use cases can be found in \cite{borgo2007foundations}, who proposed an ontology to describe a holonic organization of manufacturing and provide a classification of ADACOR (\cite{leitao2006adacor}) architecture concepts concerning the DOLCE foundational ontology. A different approach was followed by \cite{kourtis2019rule}, who focused on the requirement of a representation that enables deductive reasoning to assist human experts in decision-making in manufacturing environments.

Statistical and AI models are used in demand forecasting and production planning to support decision-making. E.g., \cite{bergman2017bayesian} describes Bayesian models' usage to improve demand forecasting of spare parts in new equipment programs. In the production planning domain, \cite{mula2006models} provide an overview of AI and simulation models for production planning under uncertainty.

To realize the simulation models in the ontology, concepts from domain-specific ontologies can be used. \cite{cannataro2003data} developed the DAMON (Data Mining Ontology for Grid Programming), which provides a reference model for data mining tasks, methodologies, and available software. \cite{panov2008ontodm,panov2014ontology} developed the Onto-DM, which provides a heavy-weight model of data mining entities, inductive queries, and data mining scenarios. More recently, \cite{sacha2018vis4ml} developed the VIS4ML ontology, focusing on the machine learning process's visual analytics aspect. \cite{diamantini2009kddonto} developed KDDONTO, an ontology for the discovery of data mining algorithms. A data mining ontology with the purpose of algorithm selection and meta-mining was developed by \cite{hilario2009data}.

From the literature review, we found no ontology was built to integrate domain knowledge regarding DTs, statistical and AI models and the two use cases proposed to aid users in decision-making in the manufacturing domain.

\subsection{Methodologies for ontology design in manufacturing}
In the previous section, we observed that we have ontologies from different domains. However, no ontology exists to integrate all relevant concepts from Demand Forecasting, Production Planning, statistical, and AI models for decision-making. Therefore, a methodology is required to develop such an ontology focusing on Demand Forecasting and Production Planning to ensure we capture all decision-making requirements in these two scenarios.

Multiple general methodologies were proposed to build an ontology. \cite{uschold1996building, uschold1995towards, fernandez1997methontology} agree that the starting point is to identify the purpose of the ontology. In contrast, \cite{staab2001knowledge} prioritizes a feasibility study to identify the problem, opportunity areas, and a potential solution. Once defined, \cite{uschold1996building,kim1995ontology} urge to define the ontology scope and emphasize the importance of deciding on the level of formality required. \cite{kim1995ontology} suggests to continue by defining a problem statement, and competency questions, to then elicit required knowledge from multiple sources, identify key concepts and relationships, and identify terms to refer to concepts and relationships (\cite{uschold1996building, uschold1995towards, fernandez1997methontology}). Finally, \cite{uschold1995towards, fernandez1997methontology} agree that an effort should be made to integrate to pre-existing ontologies and evaluate against a frame of reference.

On top of the general methodologies to construct an ontology, specific ones exist addressing the manufacturing domain. \cite{lin2011developing} proposes a two-level modeling approach combining Unified Modelling Language and Object Constrained Language, translating entities from a software object model to ontology entities. \cite{ameri2012systematic} developed a four-step methodology using a Simple Knowledge Organization System framework to develop a thesaurus of concepts, then identify relevant classes and provide logical constraints and rules. A similar approach was developed by \cite{chang2010development}, with an emphasis on manufacturing design.

Existing methodologies provide a solid foundation for concept mapping. There are many complexities in these scenarios on identifying relevant domain concepts to support the ontology, and support cognition, which is valuable in such complex systems. Therefore, systems thinking approaches can understand and model relationships, feedback loops, and causality in those complex systems.

\section{Actionable Cognitive Digital Twin}\label{ACT-DEFINITION}

\subsection{Actionable Cognitive Twin definition}

\begin{figure*}[!ht]
\centering
\includegraphics[width=5.0in]{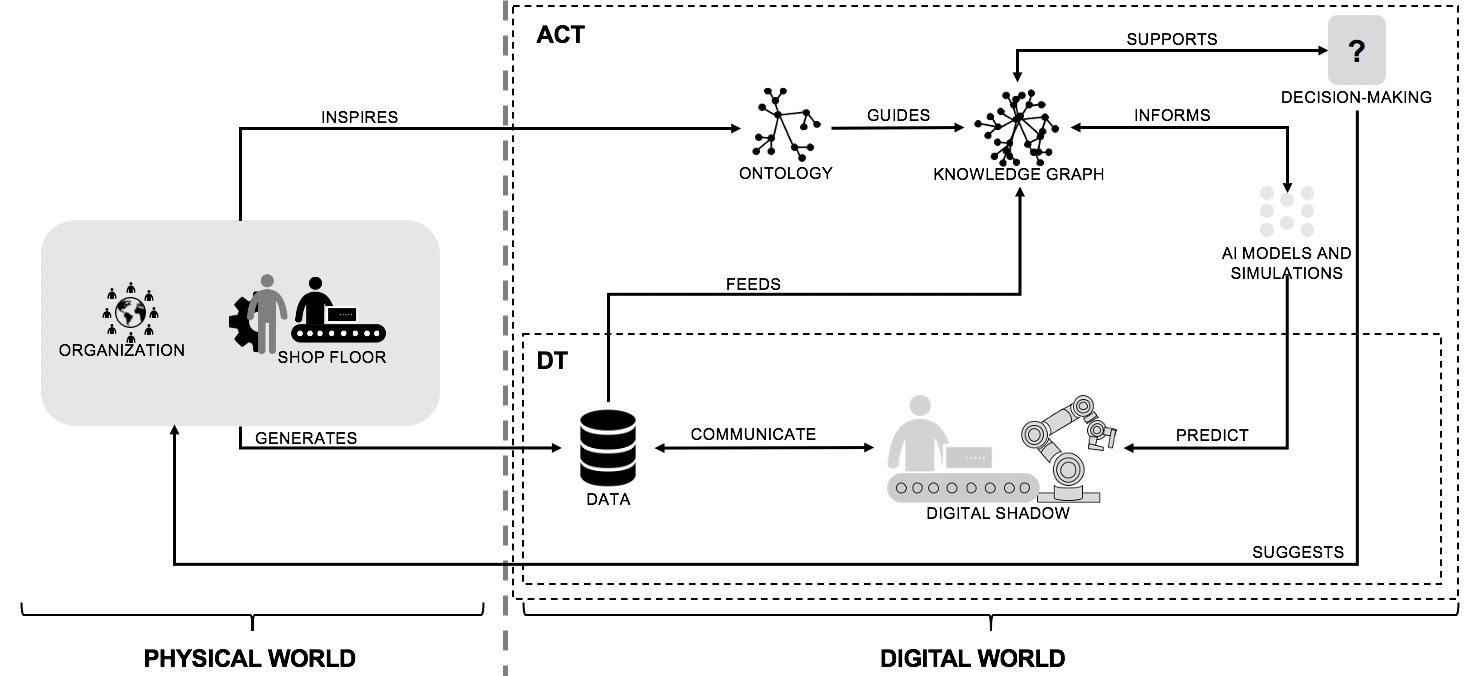}
\caption{Overview Actionable Cognitive Twins. The Actionable Cognitive Twins enhance the Digital Twins by enabling reasoning, binding different knowledge processes and types of knowledge, and providing an actionable dimension.}
\label{F:ACT-OVERVIEW}
\end{figure*}

Digital Twins formalize production processes in the digital world and get insights based on simulations and predictive analytics realized through AI and statistical models. Digital Twins with cognitive capabilities enable self-learning based on captured domain knowledge, historical data, simulations, and predictive models when implementing a production process (see Fig.~\ref{F:ACT-OVERVIEW}).

The new actionable cognitive digital twin (ACT) can self-improve through the cognitive and actionable dimensions:

\begin{enumerate}
  \item  The cognitive dimension of the Digital Twin, introduced in Section~\ref{INTRODUCTION}, is clarified based on the definition of knowledge which is described through three elements: 1) claim (manifestation of knowledge), 2) provenance (origin of the piece of knowledge), and 3) inference (a conclusion reached based on evidence and reasoning) (\cite{ullah2020knowledge}). Thus, the cognitive dimension has four types of knowledge based on such three elements, including 1) definitional (uncontroversial definition of concepts: the claim is equal to knowledge provenance, and there is no inferred knowledge), 2) deductive (knowledge obtained by deduction applied on concepts: a claim is different from knowledge provenance, where knowledge inference is made by deduction and knowledge provenance represent pieces of definitional knowledge), 3) inductive (knowledge obtained through induction applied data: a claim is different from knowledge provenance, where knowledge inference is made by induction and knowledge provenance correspond to data and/or observations), and 4) creative knowledge (knowledge due to abduction: there is no knowledge provenance, and a claim is made describing a plausible conclusion based on a priori-based creative knowledge, without positively verifying the conclusion).
  \item  An actionable dimension is required to use knowledge and insights to provide decision-making options to take action in the manufacturing plant. The actionable dimension comprehends three action types: 1) the execution of a knowledge process (run a simulation, AI model, or execute some reasoning rules), 2) autonomously execute some actions in the manufacturing plant, and 3) provide decision-making options to the user.
  \end{enumerate}

\begin{figure*}[!ht]
\centering
\includegraphics[width=5.0in]{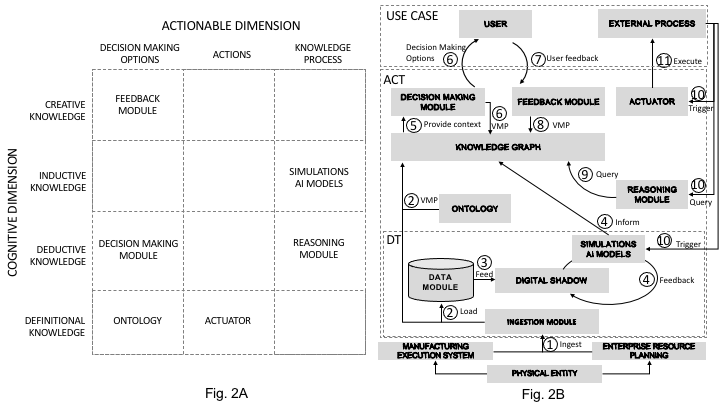}
\caption{Actionable Cognitive Twin conceptual diagrams. Fig. 2A shows architecture modules we require to realize different aspects of the cognitive and actionable dimensions. Fig. 2B provides an overview of the ACT architecture and modules interaction.}
\label{F:act-concept}
\end{figure*}

As shown in Fig.~\ref{F:act-concept}-A, the cognitive and actionable dimensions relate to each other. Definitional knowledge provides concepts and rules through an ontology implemented in a Knowledge Graph. It can be used to bind different ACT components and support reasoning. When new data or process outcomes are available, knowledge processes can be executed. These knowledge processes can involve executing reasoning rules (generating deductive knowledge), executing a simulation or AI models training (both produce inductive knowledge based on existing data). Definitional knowledge regarding possible actions to be executed or decision-making opportunities and conditions when such actions are required are used to take action or hint decision-making opportunities to the user. Feedback can be collected regarding actions taken by the user to understand which of the proposed actions or decision making-opportunities were executed in a given context. Such feedback increases the knowledge base and enables new knowledge processes to identify patterns common to those actions and decision-making opportunities. Information regarding these patterns can be used to make better decisions regarding actions or better select and rank decision-making opportunities for a given user. Such information can also inform the user and ask for additional definitional knowledge regarding actions or decision-making options that may be missing and enhance the existing knowledge base. Finally, based on experts' experience, we can also encode creative knowledge regarding plausible explanations or rules associated with particular contexts, which can be proposed to the end-users.

To realize the aforementioned cognitive and actionable capabilities, we envision different modules that are required, which collaborate, as shown in Fig.~\ref{F:act-concept}-B. The \textit{Ingestion Module} interfaces with physical entities to ingest data into the DT, loading it to the \textit{Data Module} and leveraging \textit{Virtual Mapping Procedures} (VMP), which allow enriching the incoming data with \textit{Ontology} concepts and load it to the \textit{Knowledge Graph}. The \textit{Data Module} provides the required data to create a \textit{Digital Shadow} - a digital surrogate of the \textit{Physical Entity}. In order to provide insights on the most likely future state of the \textit{Digital Shadow}, \textit{Simulations and AI models} can provide forecasts, which can enrich the \textit{Digital Shadow}, and are informed to the \textit{Knowledge Graph}. Based on those forecasts, the \textit{Decision-Making Module} can execute heuristics and identify decision-making options that could help mitigate potential issues. Those decision-making options are informed to the user, who can provide feedback on them through the \textit{Feedback Module}. The provided feedback is stored in the \textit{Knowledge Graph}. Through it, the ACT can build knowledge regarding proposed decision-making options and which are preferred for particular situations. Whenever some inferencing is required, the \textit{Reasoning Module} provides means to interface with the \textit{Knowledge Graph} and execute specific queries. The \textit{Simulations and AI models} module is tasked with providing forecasts. To realize them, it learns from past data, providing an inductive estimate of future behavior. External processes have the responsibility to trigger simulations or AI model training. An \textit{Actuator} can trigger external processes, providing an additional actionable capability to the ACT.

\subsection{Actionable Cognitive Twin architecture}
In order to realize the two dimensions described in the subsection above, we propose an ACT architecture (illustrated in Fig.~\ref{F:act-concept}) that contemplates the following components:

\begin{enumerate}
  \item \textbf{Physical entity}: entity, which has a digital counterpart in the DT. We consider physical and abstract entities such as manufacturing processes.
  \item \textbf{Digital shadow}: the digital counterpart of a physical entity.
  \item \textbf{Ontology}: capture the knowledge of interacting reference entities, their relationships, and rules related to those concepts mirrored in the digital world.
  \item \textbf{Knowledge Graph}: instantiates the ontology based on ingested data and supports arbitrary relationships between entities (\cite{paulheim2017knowledge}). Queries support access to pieces of knowledge and rules enforcement.
  \item \textbf{Ingestion module}: provides means to listen or retrieve data from different sources and map it to KG instances taking into account entity definitions and relationships provided by the reference ontology.
  \item \textbf{Data module}: stores or streams facts collected from sensors, databases, and other data sources, which describe the state of DTs reference instances, their digital counterparts, and interactions
  \item \textbf{Reasoning module}: provides means to obtain deductive knowledge by applying rules to the KG or obtain inductive knowledge from AI and statistical models. Such knowledge can be persisted to the KG.
  \item \textbf{Simulations and AI models}: learn behavior from past data in order to provide some prediction on non-observed instances
  \item \textbf{Decision Making Module}: given some meaningful event, it links it to relevant context, either by semantic or spatio-temporal proximity, or other criteria, and provides contextualized decision-making opportunities based on domain knowledge encoded in the Knowledge Graph.
  \item \textbf{Feedback module}: gathers feedback from the user, given a specific list of decision-making opportunities.
  \item \textbf{Actuator}: interfaces with some external entity or system and allows the execution of specific actions.
\end{enumerate}

\section{Ontology design and Knowledge  Graph Modeling}\label{SYSTEM-THINKING}

We propose an ontology development approach in the context of ACTs, using systems thinking. Systems thinking is used to capture ontology concept entities and their interrelationships in different use cases. Based on the ontology concepts, Basic Formal Ontology (BFO) (\cite{Smith2002}) and Industrial Ontologies Foundry (IOF) specifications (\cite{Kulvatunyou2018}) are referred to define the ontology focused on the use cases. Finally, a Knowledge Graph is developed to instantiate the ontology in the context of ACTs development.

\begin{figure*}[!t]
\centering
\includegraphics[width=5.0in]{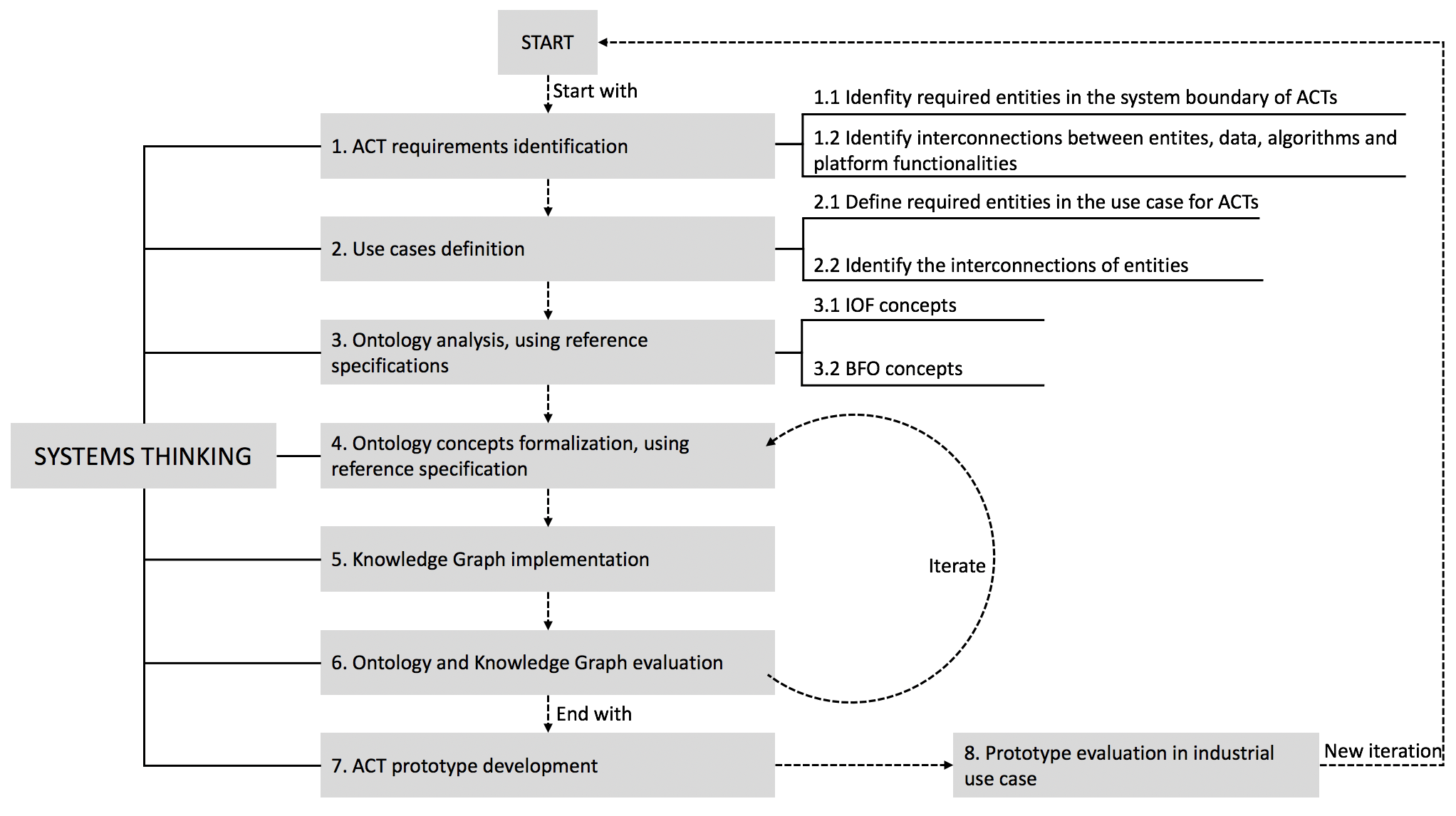}
\caption{Workflow of Systems Thinking approach to ACTs}
\label{F:ACT-systems-thinking}
\end{figure*}

\subsection{Ontology development based on systems thinking approach}

Systems thinking is an approach for understanding a system's nature by identifying the entities and their interactions within the system boundary (\cite{Haskins2014}). It aims to identify how the components are related to each other and how complex entities or processes interact with others. When using systems thinking, it is important to identify interactions between ACT components, feedback loops between users and the ACT, and pay attention to causes that trigger changes in the system, such as changes in the physical entity, new knowledge obtained from knowledge processes, or users' feedback.

We propose a system thinking approach to develop ACT ontologies and build a KG (see Fig.~\ref{F:ACT-systems-thinking}), which is comprised of several steps:
\begin{enumerate}
  \item \textbf{ACT requirements identification}: define purpose and scope of the ACT. Identify the ACT level of granularity and detail required, components and their interrelationships, and key concepts required for ontology formalisms.
  \item \textbf{Use cases definition}: define use cases and the role of the ACT in them.
  \item \textbf{Ontology analysis, using reference specifications}: analyze the IOF and BFO concepts based on the system boundary of ACTs and use cases to provide an initial ontology development plan.
  \item \textbf{Ontology concepts formalization, using reference specification}: based on IOF and BFO, define the concepts and relations for the ontology and build the ontology.
  \item \textbf{Knowledge Graph implementation}: implement a KG based on the ontology developed in the previous step and available data regarding the use cases of interest.
  \item \textbf{Ontology and Knowledge Graph evaluation}: evaluate the ontology and KG with a set of predefined metrics and criteria.
   \item \textbf{ACT prototype development}: development of ACT prototypes based on proposed ontology and KG.
  \item \textbf{Prototype evaluation in industrial use case}: evaluate the prototype on given industrial use cases. Based on evaluation outcomes, feedback, and new requirements, start a new development cycle.
\end{enumerate}

\begin{figure*}[!t]
\centering
\includegraphics[width=5.0in]{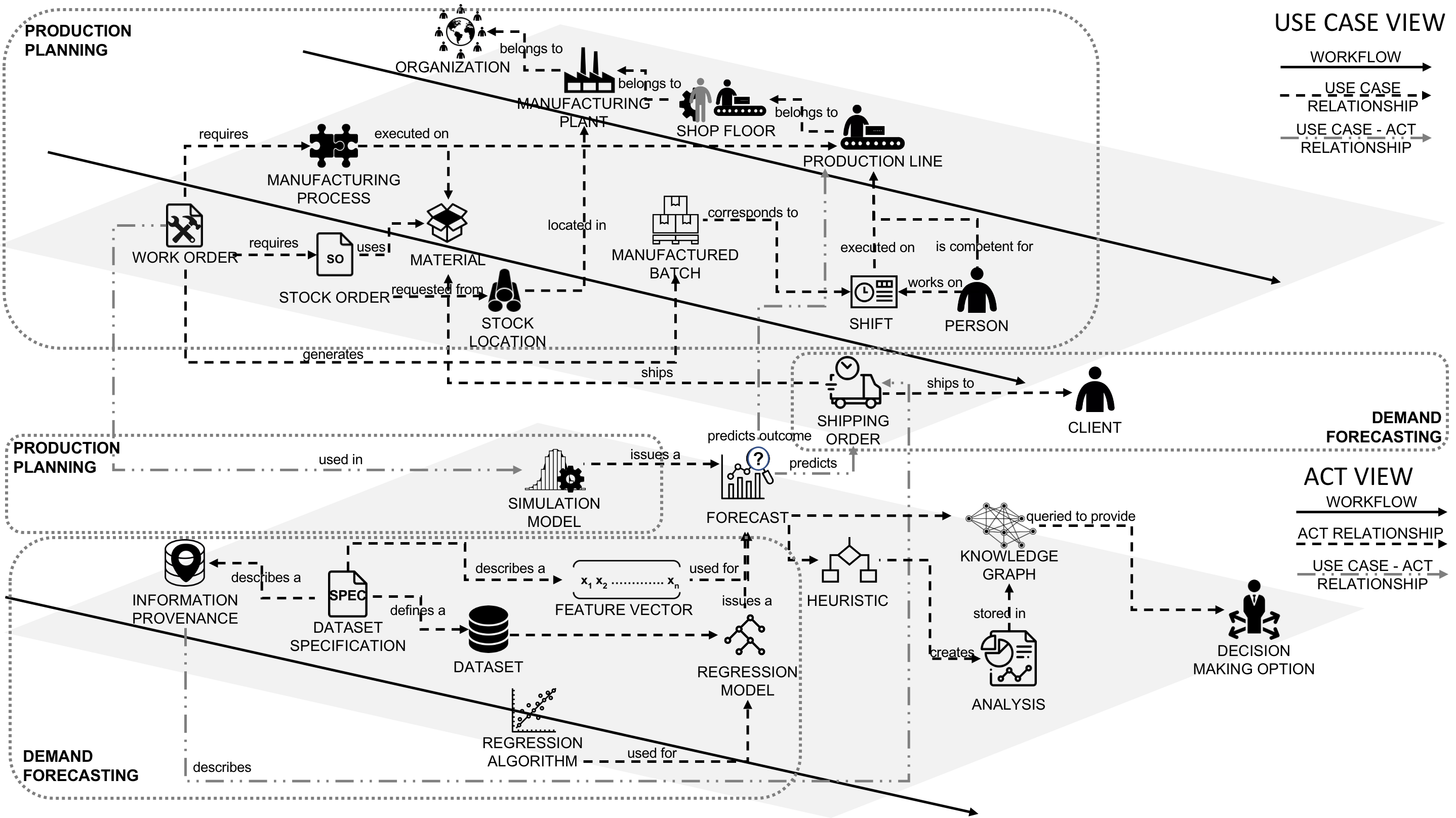}
\caption{View of the use case and ACT entities, their workflow and interactions.}
\label{F:ACT-act-and-use-case-view}
\end{figure*}

\subsection{Ontology design}

Based on the systems thinking approach presented in Section \ref{RELATED-WORK-DSO}, we develop domain-specific ontologies using a Basic Formal Ontology framework (\cite{Smith2002}), which is a top-level ontology to promote interoperability among domain ontologies built in its terms through a process of downward population. Moreover, in order to make this ontology more generally accepted, we make use of the Industrial Ontologies Foundry domain concepts in the developed ontology as references (\cite{Kulvatunyou2018}), which are proposed by the IOF community working with government, industry, academic, and standards organizations to advance data interoperability in their respective fields. 

The scenario that we developed ontology for the production planning is shown in Fig. \ref{F:ACT-act-and-use-case-view}. There are two views we introduced to describe this scenario: 1) the use case view, referring to the perspective that shows interactions between entities and stakeholders for demand forecasting and production planning; 2) the ACT view, referring to the perspective how the ACT works during the demand forecasting and production planning. The detailed views are introduced as follow:

\begin{enumerate}
     \item \textbf{Use case view}: the \textit{Manufacturing Plant} is a specific part of \textit{Organization}. \textit{Production lines} in each \textit{Manufacturing Plant} construct \textit{Shop Floors} where \textit{Manufacturing Processes} are implemented. \textit{Production Planning} is used to organize \textit{Work Orders} and \textit{Persons} in order to satisfy the production goals, while respecting legal constraints, and personal preferences. \textit{Work Orders} indicate that \textit{Manufacturing Process} should be executed, in which \textit{Shift} and \textit{Production Line} create a \textit{Manufactured Batch}. The \textit{Manufacturing Process} makes use of the required \textit{material}, which is requested through \textit{Stock Orders} based on a given \textit{Stock Location} in a \textit{Manufacturing Plant}. \textit{Persons} who work in a given \textit{Manufacturing Plant} are assigned to a \textit{Shift} and \textit{Manufacturing Line} based on their personal preferences on their shift, production goals, and legal constraints. The manufactured \textit{Material} is delivered to the \textit{Client} according to \textit{Shipping orders}, which were created based on \textit{Client} purchases.
     \textit{Demand Forecasting} is used to provide an estimate on the quantity and date materials that will be shipped to a \textit{Client}. These estimates help to make better decisions concerning material purchases and \textit{Production Planning}.
     
  \item \textbf{ACT view}: provides insight on how the ACT supports the use cases, with cognitive capabilities and  decision making options. To support \textit{production planning}, \textit{Simulation Models} are created based on the proposed \textit{Work Orders} to provide a \textit{Forecast} for production outcomes. \textit{Heuristics} produce \textit{Analysis} informing if the production goals are met or if some improvements are required to optimize operational metrics, such as organizational downtime or overall equipment efficiency. Based on \textit{Analysis} outcomes and their related domain knowledge in the \textit{Knowledge Graph}, \textit{Decision Making Options} are offered to the user. When implementing \textit{Demand Forecasting}, \textit{Shipping orders} are identified through \textit{Information Provenance}. The \textit{Information Provenance} is considered as an important references when developing \textit{Dataset Specification}, which describes how to build the \textit{Dataset}. The \textit{Dataset}, along with a \textit{Regression Algorithm}, is used to develop a \textit{Regression Model}. The \textit{Regression Model} is used to provide \textit{Forecasts} with given input \textit{Feature Vector}, which is defined based on \textit{Dataset Specification}. Based on \textit{Analysis} outcomes, and the domain knowledge about \textit{Demand Forecasting} in the \textit{Knowledge Graph}, \textit{Decision Making Options} are made to the user.
  
  \item Through these two views, production planning, and demand forecasting, are implemented to support decision-making in the ACT use case. When implementing the \textit{production planning} or \textit{demand forecasting}, the \textit{knowledge graph} is queried to understand what kind of \textit{Decision Making Options} are appropriate for the given \textit{Analysis} outcomes and the specific use case. 
\end{enumerate}

\begin{table*}[!ht]
\centering
\resizebox{\columnwidth}{!}{
\begin{tabular}{|l|l|}
\hline
\textbf{Ontology concept} & \textbf{Definition}                                                                                                                                                                                                                                                                                                                                                                         \\ \hline
Shift                     & The time period during which a person is at work.                                                                                                                                                                                                                                                                                                                                           \\ \hline
Manufacturing process     & \begin{tabular}[c]{@{}l@{}}Process in which materials are changed, converted, or transformed into a different state or form from \\ which they previously existed and includes refining materials, assembling parts, and preparing raw \\ materials and parts by mixing, measuring, blending, or otherwise committing such materials or parts \\ to the manufacturing process.\end{tabular} \\ \hline
Data source               & \begin{tabular}[c]{@{}l@{}}A data source is anything that produces digital information, from the perspective of systems which \\ consume this information.\end{tabular}                                                                                                                                                                                                                     \\ \hline
Feature vector            & Individual measurable property or characteristic of a phenomenon being observed.                                                                                                                                                                                                                                                                                                            \\ \hline
Decision-making option    & Different alternatives or solutions under consideration when making a decision.                                                                                                                                                                                                                                                                                                             \\ \hline
Use case                  & \begin{tabular}[c]{@{}l@{}}A list of actions or event steps typically defining the interactions between a role and a system to\\ achieve a goal.\end{tabular}                                                                                                                                                                                                                               \\ \hline
Forecast                  & Prediction of a future value.                                                                                                                                                                                                                                                                                                                                                               \\ \hline
Model                     & \begin{tabular}[c]{@{}l@{}}A simplified representation of a system at some particular point in time or space intended to\\ promote understanding of the real system.\end{tabular}                                                                                                                                                                                                           \\ \hline
Simulation model          & A model that behaves or operates like a given system when provided a set of controlled inputs.                                                                                                                                                                                                                                                                                              \\ \hline
Regression model          & \begin{tabular}[c]{@{}l@{}}A model that summarises relationships between an array of continuous input variables, and a \\ continuous output variable.\end{tabular}                                                                                                                                                                                                                          \\ \hline
Dataset specification     & \begin{tabular}[c]{@{}l@{}}Is a data item specification about a dataset defined with a data type specification of the data examples\\ aggregated in the dataset.\end{tabular}                                                                                                                                                                                                               \\ \hline
Dataset                   & A collection of data.                                                                                                                                                                                                                                                                                                                                                                       \\ \hline
Algorithm                 & \begin{tabular}[c]{@{}l@{}}A finite sequence of well-defined, computer-implementable instructions, to solve a class of problems\\  or to perform a computation.\end{tabular}                                                                                                                                                                                                                \\ \hline
Regression algorithm      & \begin{tabular}[c]{@{}l@{}}Regression algorithms predict the output values based on input features from the data fed in the\\ system.\end{tabular}                                                                                                                                                                                                                                          \\ \hline
Information provenance    & The place of origin or earliest known history of certain data.                                                                                                                                                                                                                                                                                                                              \\ \hline
Time series               & A series of data points indexed (or listed or graphed) in time order.                                                                                                                                                                                                                                                                                                                       \\ \hline
Work order                & A task or a job for a customer, that can be scheduled or assigned to someone.                                                                                                                                                                                                                                                                                                               \\ \hline
Stock order               & \begin{tabular}[c]{@{}l@{}}The ordering of new stock to refill the inventory, replenish shelves, or when a large order has been\\  made etc.\end{tabular}                                                                                                                                                                                                                                   \\ \hline
Shipping order            & \begin{tabular}[c]{@{}l@{}}A copy of the bill of lading containing the shipper's instructions to the carrier for transmission of \\ goods.\end{tabular}                                                                                                                                                                                                                                     \\ \hline
Stock location            & The area in the warehouse where the stock item is stored.                                                                                                                                                                                                                                                                                                                                   \\ \hline
Artifact                  & \begin{tabular}[c]{@{}l@{}}A general term for an item made or given shape by humans, such as a tool or a work of art, especially \\ an object of archaeological interest.\end{tabular}                                                                                                                                                                                                      \\ \hline
Manufactured batch        &        \begin{tabular}[c]{@{}l@{}}A group of identical or similar items that are produced together, and which go through a stage of the \\
production process before moving onto the next one to make the desired product.                                                                   \end{tabular}

\\ \hline
Material                  & \begin{tabular}[c]{@{}l@{}}An independent continuant that at all times at which it exists has some portion of matter as \\ continuant part.\end{tabular}                                                                                                                                                                                                                                    \\ \hline
Person                    & Is a member of the species Homo sapiens.                                                                                                                                                                                                                                                                                                                                                    \\ \hline
Client                    & A company that receives a service from them in return for payment.                                                                                                                                                                                                                                                                                                                \\ \hline
Shop floor                & \begin{tabular}[c]{@{}l@{}}The part of a workshop or factory where production as distinct from administrative work is carried \\ out.\end{tabular}                                                                                                                                                                                                                                          \\ \hline
Production line           & \begin{tabular}[c]{@{}l@{}}Is an artifact aggregate enabling a set of sequential operations established in a plant site where \\ components are assembled to make a finished article or where materials are put through a refining\\ process to produce an end-product that is suitable for onward consumption.\end{tabular}                                                                \\ \hline
Production plant          & \begin{tabular}[c]{@{}l@{}}An Artifact that is consisting of buildings and machinery, or more commonly a complex having \\ several buildings, where workers manufacture goods or operate machines processing one product\\ into another.\end{tabular}                                                                                                                                       \\ \hline
Organization              & \begin{tabular}[c]{@{}l@{}}An object aggregate that corresponds to social institutions such as companies, or societies, that\\ does something.\end{tabular}                                                                                                                                                                                                                               \\ \hline
\end{tabular}}
\caption{Ontology concepts for the production planning and demand forecasting use cases. \label{T:OCPP}}
\end{table*}

\begin{figure}[!t]
\centering
\includegraphics[width=5.25in]{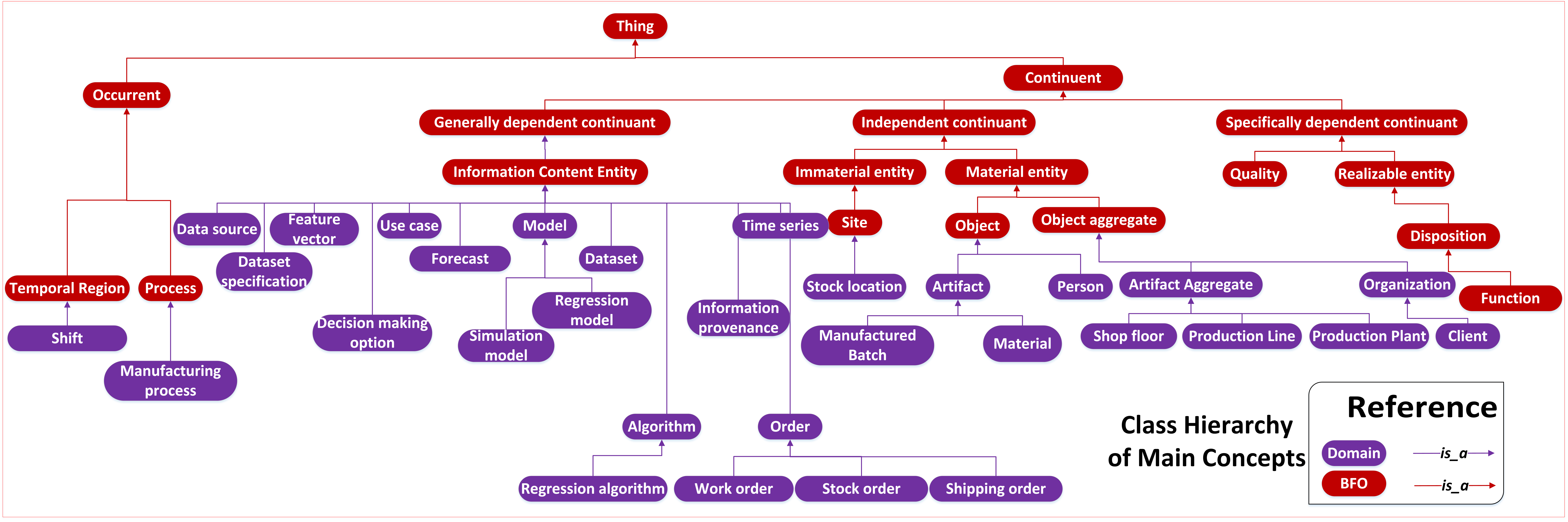}
\caption{Ontology definition based on BFO}
\label{F:ONTOLOGY-DEFINITION}
\end{figure}

Based on the BFO and IOF reference ontology, we provide a unified ontology that describes concepts related to the use case of production planning and demand forecasting. The cognitive and actionable aspects are considered throughout the whole ontology design cycle. They materialize in a set of rules applied in the reasoning module to realize deductive reasoning (Fig.~\ref{F:act-concept}) and trigger actions during decision-making. \footnote{The ontology was published and is available online (\cite{DVN/DVZH81_2021})}. Because of using upper ontologies framework BFO, the hierarchy of the upper and domain ontologies is shown in Fig.~\ref{F:ONTOLOGY-DEFINITION}. The entities of ontology concepts are defined as two types: 1) \textit{occurrent} (entities that occur) and 2) \textit{continuent} (entities that persist in time;  The details are shown in \cite{Arp2015}. Thus, the process's concepts, including the manufacturing process and prediction process, are developed under the \textit{occurrent} entity. The \textit{material entities}, such as \textit{person} are defined under \textit{continuent} entity. 

\begin{figure*}[!t]
\centering
\includegraphics[width=5.0in]{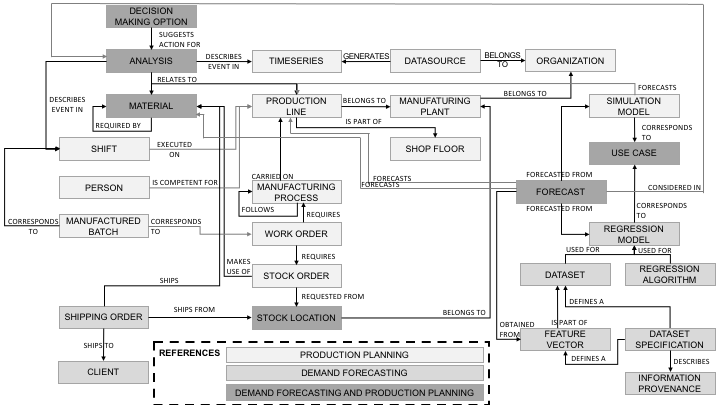}
\caption{The topology of domain ontology concepts based on BFO}
\label{F:TOPOLOGY-FOR-DOMAIN-ONTOLOGY-CONCEPTS}
\end{figure*}

Domain ontologies are developed under the BFO framework, which provides ontology entities representing the use case view and actionable view for production planning and demand forecasting, such as data source, work orders. To drive the development of the domain-specific ontology (shown in detail in Fig.~\ref{F:ONTOLOGY-DEFINITION}), we first created a topology of domain concepts (see Fig.~\ref{F:TOPOLOGY-FOR-DOMAIN-ONTOLOGY-CONCEPTS}), with concepts related to production planning and demand forecasting. The ACT obtains \textit{Timeseries data} from \textit{Datasources}, which belong to a specific \textit{Organization}. The \textit{Timeseries data} can describe the operational status of a given production line, e.g., the count of manufactured products or scrap levels observed during the manufacturing process. The \textit{Timeseries data} is analyzed with algorithms and heuristics based on \textit{Forecasts} to provide \textit{Decision Making Options} to the users when detecting anomalous situations. Based on these concepts, the production planning and demand forecasting are detailed as follow:

Production planning is concerned with planning and allocating \textit{Materials}, \textit{Persons} and \textit{Production Lines} to fulfill the required \textit{Work Orders} on time. To that end, decisions are made regarding to which \textit{Production Lines} enable to execute required \textit{Manufacturing Processes} during specific \textit{Shifts}. Scheduled \textit{Work Orders} related to \textit{Stock Orders} retrieve required \textit{Materials} from \textit{Stock Locations}. As a result, the \textit{Manufacturing Processes} are organized by a \textit{Manufactured Batch}, which is traced to a specific \textit{Work Order}. The \textit{Production Lines} belong to \textit{Manufacturing Plants} and they are arranged in a specific \textit{Shop Floor}. To optimize production planning, a \textit{Simulation Model} is required, to predict performances on the current \textit{Work Orders} based on past data.

In demand forecasting, we are interested in precisely forecasting \textit{Shipping Orders}, to ensure stocks for required manufactured \textit{Materials} do not run out. Each \textit{Shipping Order} provides the quantity of the required manufactured \textit{Material} in order that the \textit{Client} enables to understand quantity of the \textit{Material} is shipped, and the date on which the shipment must be finalised. To forecast \textit{Shipping Order}, a \textit{Regression Model} is built based on a \textit{Regression Algorithm} through training \textit{Dataset}. The \textit{Dataset} is defined based on a \textit{Dataset Specification}, which describes the descriptions of \textit{Information Provenance} (such as providing information where the data regarding shipping orders is stored) and data features are required to capture. The \textit{Dataset Specification} also provides insight on how a \textit{Feature Vector} is created which enables to issue a new \textit{Forecast} based on the related \textit{Regression Model}. 

\subsection{Knowledge Graph Implementation}

\begin{figure*}[!t]
\centering
\includegraphics[width=5.0in]{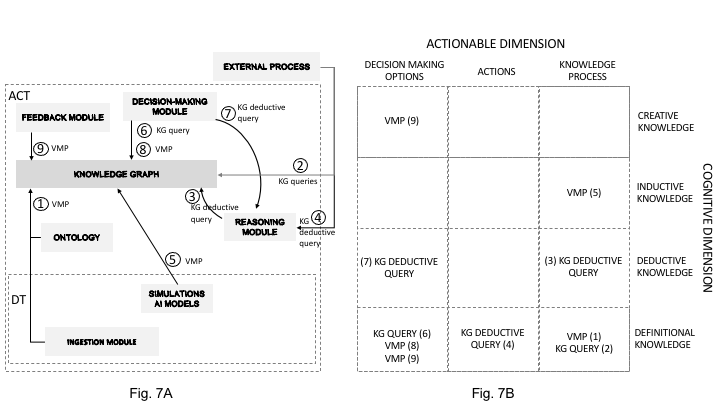}
\caption{Knowledge Graph interactions in the ACT. Fig. 7A displays Knowledge Graph interactions with other ACT components, while Fig. 7B categorizes those interactions within the actionable and cognitive dimensions.}
\label{F:ACT-kg-interactions}
\end{figure*}

Based on the defined ontology, the Knowledge Graph is modeled through Neo4j\footnote{https://neo4j.com/}. The Knowledge Graph is one of the key components of the ACT. It is used to define and process domain-specific knowledge regarding physical and ACT entities. Fig.~\ref{F:act-concept} demonstrates the content of the knowledge graph defined mostly by the virtual mapping procedures that provide concept enrichment to data from the Ingestion Module. The Reasoning Module interacts with the Knowledge Graph to extract required information for creating deductive knowledge. Outcomes of simulation and AI models are informed to the Knowledge Graph after the forecasting is done. Finally, the Decision-Making Module uses the Knowledge Graph to provide decision-making options that match certain use-case and criteria. 
 
As shown in Fig.~\ref{F:ACT-kg-interactions}, forecasts enable deductive reasoning and expose the defined knowledge to other ACT modules through a query engine. The query engine is a piece of software provided on top of the database (in our case, the Neo4j graph database) that enables us to execute queries against it. The Knowledge Graph does not only provide value on the cognitive dimension but also supports the actionable dimension by interacting with other services as shown in Fig.~\ref{F:ACT-kg-interactions}-B. We describe and exemplify these cognitive processes in Table~\ref{T:KG-INTERACTIONS}. For example, in Fig.~\ref{F:ACT-kg-interactions}-B VMP(1) corresponds to virtual mapping procedures (introduced in Section~\ref{ACT-DEFINITION}) used to load definitional knowledge by augmenting the data obtained from the Ingestion Module with semantic meaning. The VMP corresponds to a knowledge process that enables to load of the data into the knowledge graph from the actionable dimension. Such data corresponds to definitional knowledge from the cognitive dimension since it complies with semantic definitions presented in an ontology and describes entities from the physical world.

\begin{table*}[!t]
\centering
\scalebox{0.60}{
\begin{tabular}{|l|l|l|l|}
\hline
\textbf{\#} & \textbf{Knowledge Process}  & \textbf{Description} & \textbf{Example} \\ \hline
\textbf{1}  & \textbf{VMP}                & \begin{tabular}[c]{@{}l@{}}Load definitional knowledge, providing \\ semantic meaning to data obtained \\ from external sources (through the \\ ingestion module).\end{tabular} & Load properties describing a production line. \\ \hline
\textbf{2}  & \textbf{KG query}           & \begin{tabular}[c]{@{}l@{}}Queries issued by some external process \\ against the Knowledge Graph to obtain \\ definitional knowledge.\end{tabular} & The characteristics of certain production line. \\ \hline
\textbf{3}  & \textbf{KG deductive query} & \begin{tabular}[c]{@{}l@{}}Queries issued by the Reasoning Module\\ to perform deductive inferencing,  and\\ make explicit some relationships present\\ in the Knowledge Graph.\end{tabular}         & \begin{tabular}[c]{@{}l@{}}Infer that all workers associated to given \\ production lines work in a manufacturing plant, \\ and make that relation explicit.\end{tabular} \\ \hline
\textbf{4}  & \textbf{KG deductive query} & \begin{tabular}[c]{@{}l@{}}An external process interacts with the\\ Reasoning Module to perform some \\ deductive inference. The result is used to\\ execute an action.\end{tabular}                 & \begin{tabular}[c]{@{}l@{}}Find which use cases have stale simulation or \\ AI models, to update them.\end{tabular} \\ \hline
\textbf{5}  & \textbf{VMP}                & \begin{tabular}[c]{@{}l@{}}Load inductive knowledge acquired in the\\ Simulations and AI models module into \\ the Knowledge Graph.\end{tabular} & \begin{tabular}[c]{@{}l@{}}Load simulation forecasts for production lines \\into the Knowledge Graph.\end{tabular} 
\\ \hline
\textbf{6}  & \textbf{KG query}           & \begin{tabular}[c]{@{}l@{}}Queries issued by the Decision-Making\\ module against the Knowledge Graph to\\ get the required definitional knowledge.\end{tabular} & \begin{tabular}[c]{@{}l@{}}Information regarding some workers, which is\\ relevant to heuristics used to issue \\ decision-making options.\end{tabular}
\\ \hline
\textbf{7}  & \textbf{KG deductive query} & \begin{tabular}[c]{@{}l@{}}Interactions between the Decision-Making\\ module and the Reasoning module, to \\ perform some deductive inference.\end{tabular} & \begin{tabular}[c]{@{}l@{}}Infer existing decision-making opportunities\\ related to Production Planning use case that can\\ be applied to a certain production line.\end{tabular} \\ \hline
\textbf{8}  & \textbf{VMP}                & \begin{tabular}[c]{@{}l@{}}Persist the decision-making opportunities \\ advised for a particular scenario into the \\ Knowledge Graph.\end{tabular} & \begin{tabular}[c]{@{}l@{}}If we expect an organizational downtime due to\\ a missing worker under the current production\\ schedule, the module can advise either updating\\ the production schedule or add an additional shift.\end{tabular} \\ \hline
\textbf{9} & \textbf{VMP} & \begin{tabular}[c]{@{}l@{}}Persist feedback obtained from the user\\ regarding decision-making opportunities\\ advised for a given scenario. The feedback\\ can contain definitional knowledge\\ (inform what decision was taken), and \\ creative knowledge (inform a \\ decision-making option that is not \\ registered in the Knowledge Graph).\end{tabular} & \begin{tabular}[c]{@{}l@{}}The user offers feedback on the decision taken,\\ providing valuable information to rank and filter\\ decision-making opportunities in future scenarios.\end{tabular} \\ \hline
\end{tabular}}
\caption{Description of cognitive processes presented in Fig.~\ref{F:ACT-kg-interactions}-B. \label{T:KG-INTERACTIONS}}
\end{table*}

\section{Case Study}\label{USE-CASES}

The case study to evaluate our concepts is from a partner in the EU Horizon 2020 FACTLOG project\footnote{https://www.factlog.eu/}. The digital shadow is realized through a software platform, based on the data from their SAP\footnote{https://www.sap.com/} Enterprise Resource Planning software and Manufacturing Execution Systems. While the Enterprise Resource Planning software provides insights into basic manufacturing plant schedules (e.g., production, material use, and shipping, among others), the Manufacturing Execution Systems provide real-time insights regarding the manufacturing plant activities. The digital shadow describes real-world assets and processes in detail. For Production Planning, the software platform runs simulations to provide a probabilistic forecast of expected termination dates and times for scheduled work orders. These forecasts are then used to create analysis, provide insights to the users, and recommend decision-making options. For Demand Forecasting, the software regularly retrieves data from SAP, provides insights into current sales and shipping orders, and provides demand forecasts to assist planners in decision-making.

The software platform can create a Digital Twin of a manufacturing plant and its entities, providing digital shadow, simulation, and AI models (simulations for Production Planning and AI models for Demand Forecasting). Our ACT implementation enhances the existing Digital Twin implementation by providing (i) an ontology, (ii) virtual mapping procedures to load data into a knowledge graph depending on the ontology concepts, (iii) a knowledge graph, and an (iv) reasoning module. While the software platform has some heuristics to support production planning, we developed a decision-making module for demand forecasting. The aforementioned modules enable the software platform to use deductive reasoning and relate existing definitional and inductive knowledge. Through the module we proposed, the ability to perform deductive inference provides the ACT awareness of components related to a certain use case.

\begin{figure*}[!t]
\centering
\includegraphics[width=5.0in]{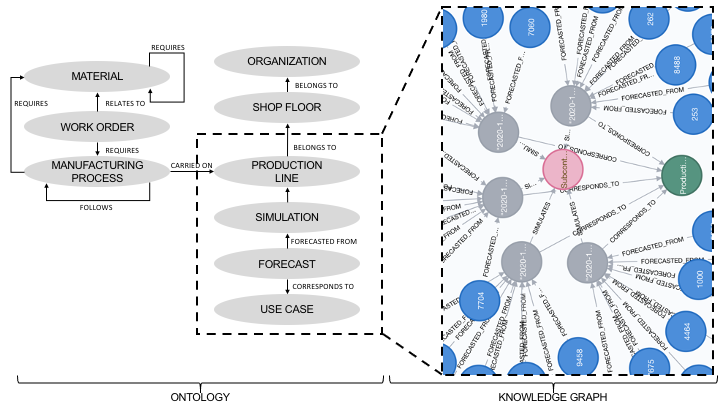}
\caption{On the left side of the Figure, we present a part of the ontology. The dotted square zooms part of the ontology, displaying nodes from our Neo4j Knowledge Graph implementation on the left.}
\label{F:ACT-kg-snapshot}
\end{figure*}

Based on the domain-specific ontology we developed, we created a Knowledge Graph using the Neo4j graph database as shown in Fig.~\ref{F:ACT-kg-snapshot}. The Knowledge Graph was instantiated with data obtained from the software platform and the demand forecasting software. The data is loaded to the knowledge graph through a virtual mapping procedure implemented as CQL ingestion scripts. Finally, queries are developed to enable deductive reasoning based on the reference of \cite{kourtis2019rule}. Two different use cases are introduced to evaluate the ACT concepts. 

\subsection{Production Planning}

Before using the software platform, production planning was handled by each planner in separate spreadsheets. Being a time-consuming task, production plans could not be manually updated when a new change is observed in the production lines. However, the software platform provides the means to regularly run simulations and determine how production processes are affected to update the production plans. Though the software platform implements a Digital Twin and provides basic decision-making options to planners, the simulation model outcomes and decision-making options are not semantically related. This leads to that Digital Twin and simulation results cannot be identified with their related use cases. Though decision-making can be realized with some heuristics, no means are provided to link them to a use case and enhance and filter them based on domain knowledge.

The ACT is used to support Production Planning with cognitive capabilities and decision-making options to solve this problem. Information regarding the work orders is retrieved from the SAP software and ingested to the software platform, which creates a \textit{Digital Shadow} of relevant entities. \textit{Simulations and AI models} module regularly runs simulations that take past orders to provide a probabilistic forecast on expected finalization dates for the scheduled orders. Heuristics are applied to these probabilistic forecasts to detect potential issues, such as organizational downtimes. Based on those outcomes, \textit{Decision Making Options} are retrieved from the \textit{Knowledge Graph}, and recommended to the user. 

Through ACT implementations, some domain problems are solved for the FACTLOG project. 
\begin{itemize}
    \item What use cases are supported with forecasting models (simulations or AI models)?
    \item How is a specific production line defined? (definitional knowledge)
    \item Do we have forecasts for the Production Planning use case? (access inductive knowledge through deductive inference)
    \item Do we have Decision Making Options for a given production line and shift? (actionable aspect exposed through deductive inference)
\end{itemize}

\begin{figure*}[!t]
\centering
\includegraphics[width=5.0in]{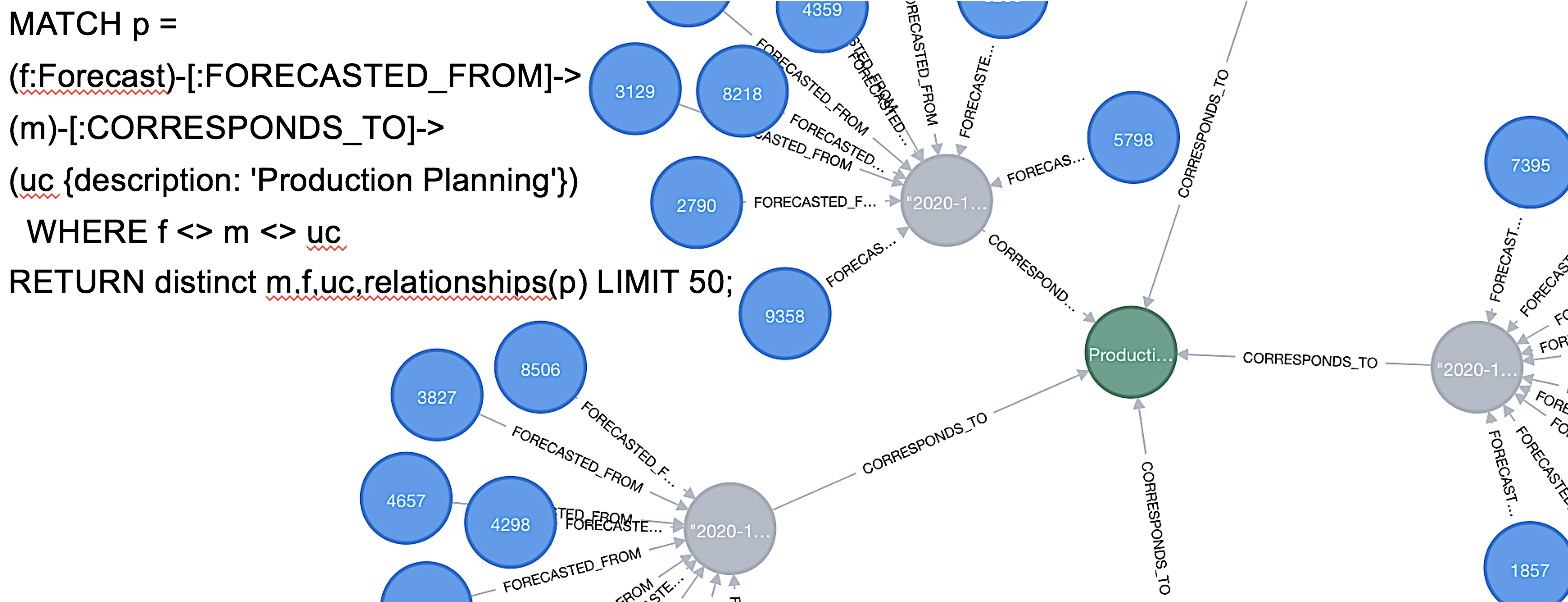}
\caption{On the left side of the Figure, we present a query we issued for the production planning use case, and on the right, the result we obtained in Neo4j. The query aims to find forecasts issued for the production planning use case.}
\label{F:ACT-query-PP}
\end{figure*}

Some knowledge graph queries are used to answer such proposed questions as shown in Listing~\ref{L:CQLQUERIES-PP}. The Knowledge Graph returns results for each of the queries listed above. We provide an example in Fig.~\ref{F:ACT-query-PP}. Though we cannot provide the results for all the queries, we published the scripts to create the Knowledge Graph and a synthetic dataset to populate it. This setting can be used to reproduce the listed queries.

\begin{lstlisting}[caption={Queries highlighting ACT capabilities for the production planning use case},label={L:CQLQUERIES-PP}]
// What use cases are supported with forecasting models?
MATCH p = (f:Forecast)-[:FORECASTED_FROM]->
(m)-[:CORRESPONDS_TO]->(uc)
WHERE f <> m <> uc
RETURN distinct m,f,uc,relationships(p);

// How is a specific production line defined? 
MATCH (pl:ProductionLine{uuid:'0a1e'}) RETURN pl;

// Do we have forecasts for Production Planning? 
MATCH p = (f:Forecast)-[:FORECASTED_FROM]->(m)
-[:CORRESPONDS_TO]->(uc {description: 'Production Planning'})
WHERE f <> m <> uc
RETURN distinct m,f,uc,relationships(p);

// Do we have Decision Making Options for a given Production Line and Shift?
MATCH p=((n:DecisionMakingOption)-[*]-(Shift{uuid:'dab85031f7414e15b6917b7d83d884e5'})-[:EXECUTED_ON]->(pl:ProductionLine{uuid:'93216b15b0b74712bcb62c0397da394e'})) RETURN p;

\end{lstlisting}

\subsection{Demand Forecasting}

Demand forecasting is usually handled by logisticians responsible for forecasting demand for a given group of products and clients. They compute forecasts manually, regularly, for some predetermined time-horizon. To compute them, they use information regarding past demand, open sales orders, and insights regarding the market growth, which they obtain from indexes, such as the Purchase Managers' Index or market analyses. Since the task is time-consuming, forecasts are not provided for many time horizons or aggregation levels. The demand forecasting module at the software platform aims to provide timely, accurate, multi-horizon demand forecasts at different aggregation levels.

The ACT supports Demand Forecasting with cognitive capabilities. Information regarding the shipping orders is retrieved from SAP software and ingested to the demand forecasting module to create a \textit{Digital Shadow} of past and current demand. The \textit{Simulations and AI models} module consumes the shipping orders to train an AI model for demand forecasting. The AI model provides the ability to predict daily demand for a given manufactured material and target client. The models are trained to provide forecasts at different time horizons, and their forecasts are aggregated at different levels. They issue state-of-the-art results providing accurate estimates of expected material shipments. For this research, we considered forecasts on a daily level. Based on forecast outcomes, the \textit{Knowledge Graph} is queried for matching \textit{Decision Making Options}. The \textit{Decision Making Options} are enriched with specific data from the forecast and advised to the user.

\begin{figure*}[!t]
\centering
\includegraphics[width=5.0in]{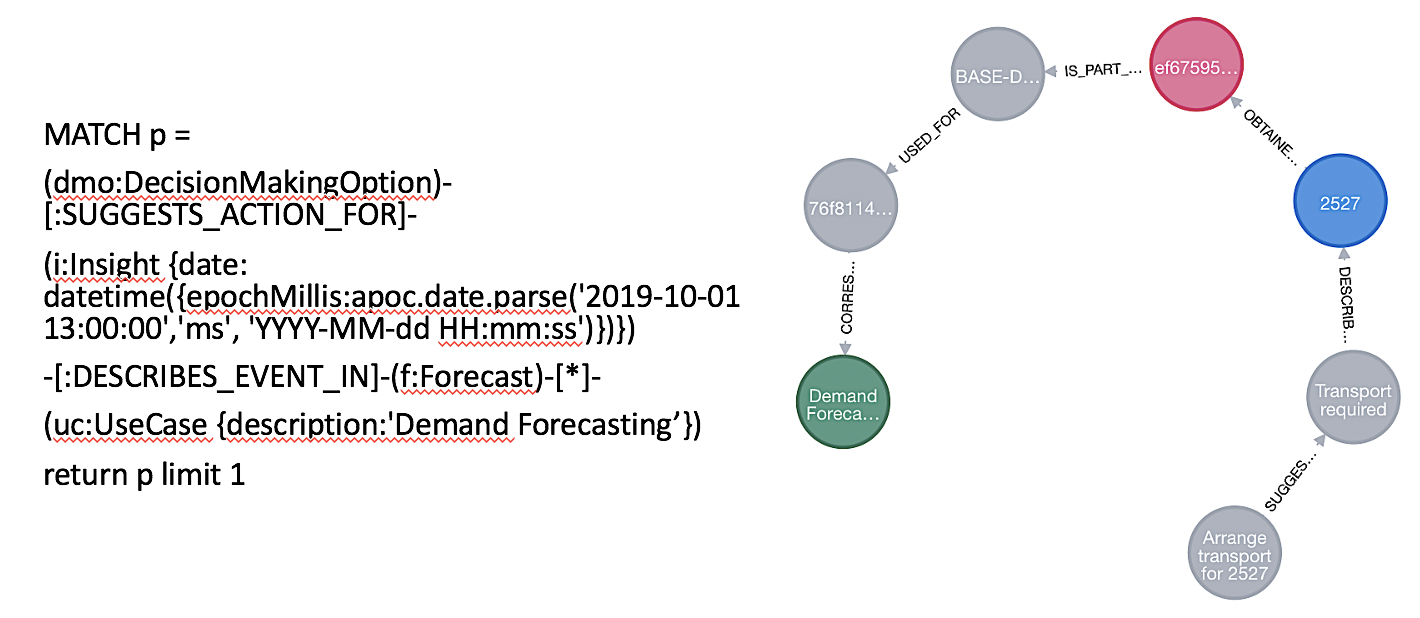}
\caption{On the left side of the Figure, we present a query we issued for the demand forecasting use case, and on the right, the result we obtained in Neo4j. The query aims to find a decision-making option for the demand forecasting use case.}
\label{F:ACT-query-DF}
\end{figure*}

We assess the usefulness of the ACT for the demand forecasting use case through two example questions and the corresponding queries we issue against the Knowledge Graph (see Listing~\ref{L:CQLQUERIES-DF}):
\begin{itemize}
    \item Do we have a forecast for material shipments for a given date?
    \item What decision-making options are suggested for the Demand Forecasting use case for a given date?
    \item What use cases relate to shipment forecasts?
\end{itemize}
We provide an example in Fig.~\ref{F:ACT-query-DF}. Though we cannot provide the results for all the queries, we published the scripts to create the Knowledge Graph and a synthetic dataset to populate it. This setting can be used to reproduce the listed queries.

\begin{lstlisting}[caption={Queries highlighting ACT capabilities for the demand forecasting use case},label={L:CQLQUERIES-DF}]
// Do we have a forecast for material shipments for a given date?
MATCH p1 = (f1:Forecast)-[:FORECASTED_FROM]->(mod1)-[:CORRESPONDS_TO]->(uc {description: 'Demand Forecasting'})
MATCH p2 = (mod2{target_date:datetime({epochMillis:apoc.date.parse('2019-07-01','ms', 'YYYY-MM-dd')})  })-[:CORRESPONDS_TO]->(uc {description: 'Demand Forecasting'})
MATCH p3 = (f2:Forecast)-[*]-(mat:Material)
WHERE f1 <> mod1 <> mat <> uc AND f1=f2 AND mod1=mod2
RETURN distinct 
mod1,f1,uc,mat,relationships(p1),relationships(p2),relationships(p3);

// What decision-making options are suggested for the 
// Demand Forecasting use case for a given date?
MATCH p = 
(dmo:DecisionMakingOption)-[:SUGGESTS_ACTION_FOR]-(i:Insight {date: datetime({epochMillis:apoc.date.parse('2019-10-01 13:00:00','ms', 'YYYY-MM-dd HH:mm:ss')})})-[:DESCRIBES_EVENT_IN]-(f:Forecast)-[*]-(uc:UseCase {description:'Demand Forecasting'}) 
return p;

// What use cases relate to shipment forecasts?
MATCH p = (uc:UseCase)-[*]-(do:ShippingOrder) return uc;
\end{lstlisting}

\section{Evaluation}\label{RESULTS}

In this research, we performed a qualitative assessment of the ACT through examples and queries provided in Section~\ref{USE-CASES}. We demonstrate that the ACT is capable of definitional, inductive, and deductive knowledge. The Knowledge Graph enables to infer implicit information contained in entities' relationships. Such inferences enable us to get information for different use cases without detailed knowledge of underlying components (e.g., type of forecasting models or decision-making options): the components can be inferred through their relationships to a given use case, or other entities.

In addition to the aforementioned results, we also published the ontology we used to model the Knowledge Graph (\cite{DVN/DVZH81_2021}).

\begin{figure*}[!t]
\centering
\includegraphics[width=4.0in]{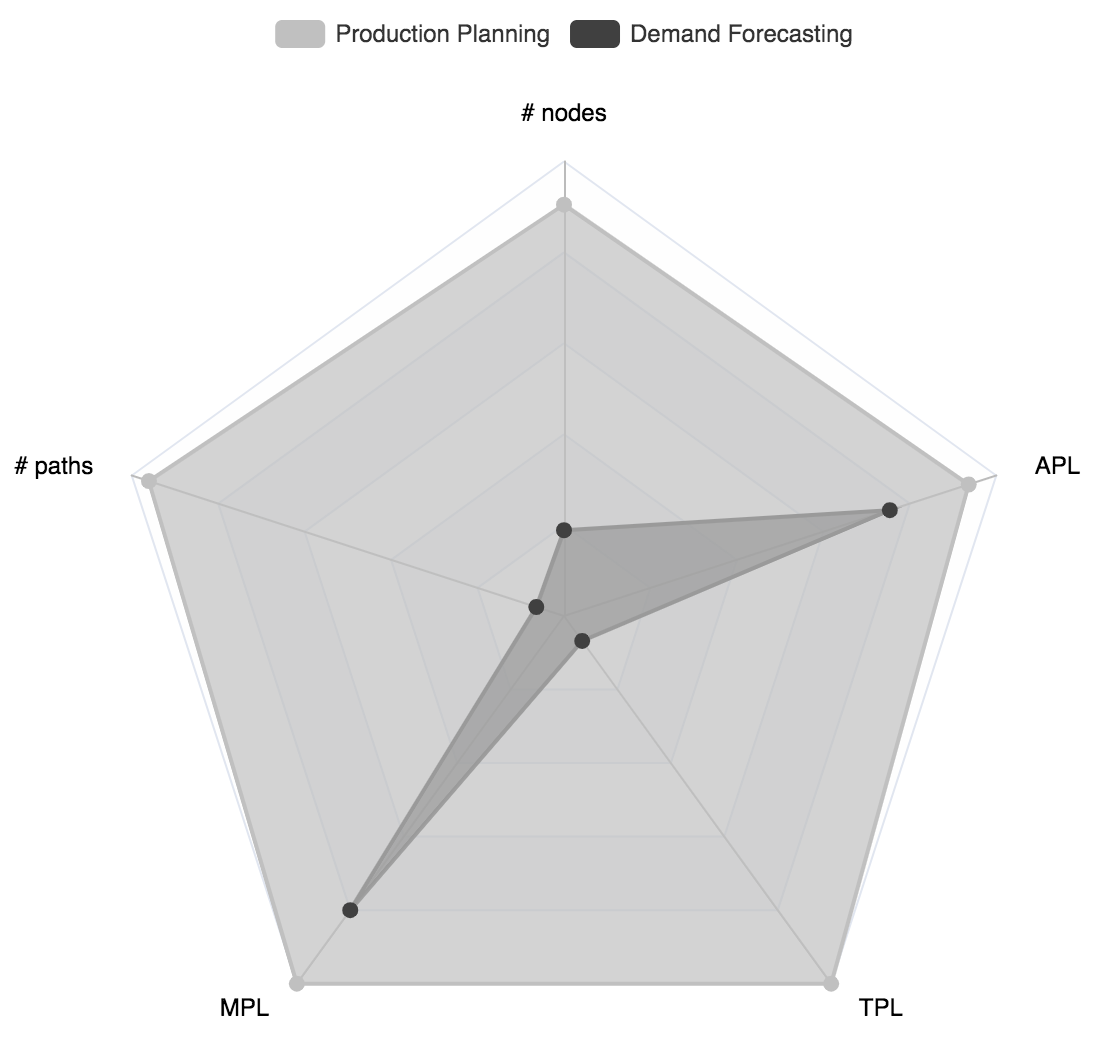}
\caption{Radar chart comparing graph metrics obtained when considering sampled nodes and relationships of the Production Planning and Demand Forecasting use cases.}
\label{F:ACT-radar-graph}
\end{figure*}

In order to assess the Knowledge Graph structure, we considered five metrics (\cite{mathieson2010complexity}): number of nodes (\# nodes), number of relationships (\# paths), total path length (TPL - the sum of relationships traversed while traveling from each node to every other node), maximum path length (MPL - maximum length among shortest paths between nodes) and average path length (APL - the average of shortest path lengths between nodes). Due to the time complexity (\cite{hartmanis1965computational}) of assessing the metrics against all nodes, we computed them on a random sample of 0.01\%. We provide the results, by use case and for the whole graph, in Table~\ref{T:SUMMARY-STATS}. We also compare the metrics between use cases in Fig.~\ref{F:ACT-radar-graph}. The Figure~\ref{F:ACT-radar-graph} shows Production Planning involves more instances than the Demand Forecasting use case. We consider this outcome also reflects the amount of domain ontology concepts related to each use case, as shown in Figure~\ref{F:TOPOLOGY-FOR-DOMAIN-ONTOLOGY-CONCEPTS}.

\begin{table*}[!t]
\centering
\scalebox{0.75}{
\begin{tabular}{|l|r|r|r|r|r|r|}
\hline
\textbf{}                    & \textbf{\# nodes} & \textbf{\# paths} &  \textbf{MPL} & \textbf{TPL} & \textbf{APL} \\ \hline
\textbf{Production Planning} & 647932            & 1650318           &  10           & 394556076    & 7,49         \\ \hline
\textbf{Demand Forecasting}  & 135134            & 110040            & 8            & 26772524     & 6,03         \\ \hline
\textbf{ALL}                 & 715985            & 1718585           & 8            & 52889866     & 5,72         \\ \hline
\end{tabular}}
\caption{Summary statistics characterizing the nodes and relationships of the knowledge graph. \# nodes and \# paths are computed on the whole graph, while MPL, TPL, and APL are calculated for a random sample of 0.01\% due to time complexity. We present statistics by use case and an overall sample. \label{T:SUMMARY-STATS}}
\end{table*}

In Section~\ref{USE-CASES} we have shown the ACT provides means to link different types of knowledge and means to associate them to a specific use case by Knowledge Graph. In Table~\ref{T:SUMMARY-STATS} we provide summary statistics of our Knowledge Graph implementation to show that such functionality is supported for a complex scenario of two use cases (Demand Forecasting and Production Planning).

\subsection{Limitations and future work}

The Knowledge Graph was developed and tested using real data provided by partners from the Europe H2020 FACTLOG project, it was not yet deployed into production environments. Given the wide range of entities modeled in the Knowledge Graph, and the results obtained for Production Planning and Demand Forecasting, the ACT implementation can be scaled to support more complex use cases.

The integration of forecast explainability remains a challenge and will be the subject of future work. The ability to provide explanations for inductive forecasts is a topic of increasing importance. While much research is being done on techniques that enable providing such explanations, much less work was done on assessing how these explanations impact users' decisions and how users' feedback can enhance them\cite{van2021evaluating}. We envision the ACT \textit{Feedback module} can be used for this purpose. Our current implementation lacks the ability to receive feedback from the users. The \textit{Feedback module} can be used to complete the \textit{Knowledge Graph} with context data that is not available through Manufacturing Execution System or Enterprise Resource Planning software by asking the user questions when anomalies are detected. Such context can be valuable to make decisions on which \textit{Decision Making Options}. We consider the feedback provided to the \textit{Feedback module} on suggested decision-making options to assess the quality of the decision-making options, create inductive recommendations, and improve their future selection and ranking.

\section{Conclusion}\label{CONCLUSION}

Actionable Cognitive Twins are made possible with an increasing level of digitalization of manufacturing processes. This paper presents a systems-thinking methodology to build Actionable Cognitive Twins, as an enhanced version of the Digital Twins, which have cognitive and actionable capabilities. Cognitive capabilities are defined based on the ability to produce knowledge, including definitional, deductive, and inductive knowledge. The actionable capability is related to executing a knowledge process, autonomously executing actions in the manufacturing plant, or suggesting decision-making options to the users. When considering the use cases, we limited our research to providing decision-making options to the users.

We illustrated our approach by implementing and evaluating the Actionable Cognitive Twin through two use cases (demand forecasting and production planning) with a European original equipment manufacturer's data. The developed domain-specific ontology and knowledge graph support decision-making in the use cases through queries. This demonstrates how the Actionable Cognitive Twin can relate different types of knowledge, benefit from the semantic representation and deductive reasoning capabilities, and suggest decision-making options for the user.

\section*{Funding}
This work was supported by the Slovenian Research Agency and the European Union’s Horizon 2020 program project FACTLOG under grant agreement H2020-869951.

\bibliographystyle{tfcad}
\bibliography{main}

\end{document}